\documentclass{article}
\PassOptionsToPackage{numbers, compress}{natbib}

\usepackage[preprint]{neurips_2026}
\usepackage{wrapfig}

\usepackage[utf8]{inputenc} 
\usepackage[T1]{fontenc}    
\usepackage{hyperref}       
\usepackage{url}            
\usepackage{booktabs}       
\usepackage{amsfonts}       
\usepackage{nicefrac}       
\usepackage{microtype}      
\usepackage{xcolor}         
\usepackage{amsmath}
\usepackage{amssymb}
\usepackage{mathtools}
\usepackage{amsthm}
\usepackage{graphicx}
\usepackage{subcaption}
\usepackage{xspace}
\usepackage{siunitx}
\usepackage{tabularx}
\usepackage{array}
\usepackage[capitalize,noabbrev]{cleveref}
\usepackage{adjustbox}
\usepackage{caption}
\usepackage{enumitem}

\usepackage{multirow}

\title{Receiver-Resolution APS--PDP Maps for Wireless Channels: Benchmarking and Coarse-to-Fine Modeling}

\author{%
  Peilin Zheng\textsuperscript{1} \quad
  Haozhe Jia\textsuperscript{2} \quad
  Xiucheng Wang\textsuperscript{1} \\
  Wenshuo Chen\textsuperscript{2} \quad
  Nan Cheng\textsuperscript{1} \\[0.6em]
  \textsuperscript{1}Xidian University \\
  \textsuperscript{2}The Hong Kong University of Science and Technology (Guangzhou)
}

\newcommand{\dataset}{UrbanADPSMap\xspace}
\newcommand{\method}{SpecFlowAR\xspace}

\begin{document}

\maketitle

\begin{abstract}
Angle Power Spectrum (APS) and Power Delay Profile (PDP) are receiver-facing spectra that govern beam selection and wideband equalization. 
Most ray-tracing benchmarks provide supervision as variable-size multipath component lists. 
This format does not match practical receivers, whose angular and delay resolutions are limited by array aperture and system bandwidth. 
As a result, learning targets are ambiguous and evaluations can change with reconstruction conventions. 
We introduce \dataset, a benchmark that converts ray-traced supervision into fixed-grid APS and PDP targets by explicitly modeling bandwidth-limited and aperture-limited response kernels. 
By fixing the receiver-domain kernels, discretization, and validity mask, \dataset removes reconstruction-induced target ambiguity and stabilizes downstream evaluation across methods. 
Built on this benchmark, we propose \method, a coarse-to-fine generator based on multi-scale autoregressive flow matching. 
The model first captures coarse dispersion structure and then refines beam-relevant peaks and low-power tails. 
The number of function evaluations controls the sampling budget, enabling latency and quality to be adjusted for dense spectrum queries.
Experiments on the test set show that \method reduces BPL@5 to 2.03 dB under a unified \(\mathrm{NFE}=80\) budget, while improving beam utility and reducing errors of 3GPP dispersion statistics.
\end{abstract}

\section{Introduction}
\label{sec:introduction}

Wideband links and highly directional transmission are becoming standard in modern wireless systems \cite{Rappaport_2013, Barati_2016}. In these regimes, link quality is governed less by a single average quantity such as path loss, and more by how received power is distributed over angle and delay at the finite resolution of practical arrays and waveforms \cite{Sayeed_2002}. The angle power spectrum (APS) drives beam search and tracking, while the power delay profile (PDP) governs equalization and delay dispersion. These receiver-facing spectra also define the moment statistics used in 3GPP evaluation, including delay spread (DS) and azimuth spread of arrival (ASA) \cite{3gpp_tr_38901}.

Despite their role in receiver operation, most ray-tracing benchmarks provide supervision as variable-size multipath component (MPC) lists \cite{Falade_2023,Alkhateeb_2019_DeepMIMO,Alrabeiah_2019_ViWi,Hoydis_2023,Taylor_2013,Gotszald_2015}. A practical receiver does not directly observe such lists. It observes spectra after finite-aperture beamforming and finite-bandwidth filtering. Converting MPCs into APS and PDP therefore requires explicit choices of response kernels and discretization rules. When these choices are implicit or inconsistent across works, both learning targets and downstream metrics can drift. Reported gains may then reflect reconstruction conventions rather than improved modeling.

\newpage
\noindent
\begin{minipage}{\linewidth}
  \centering
  \includegraphics[width=\linewidth]{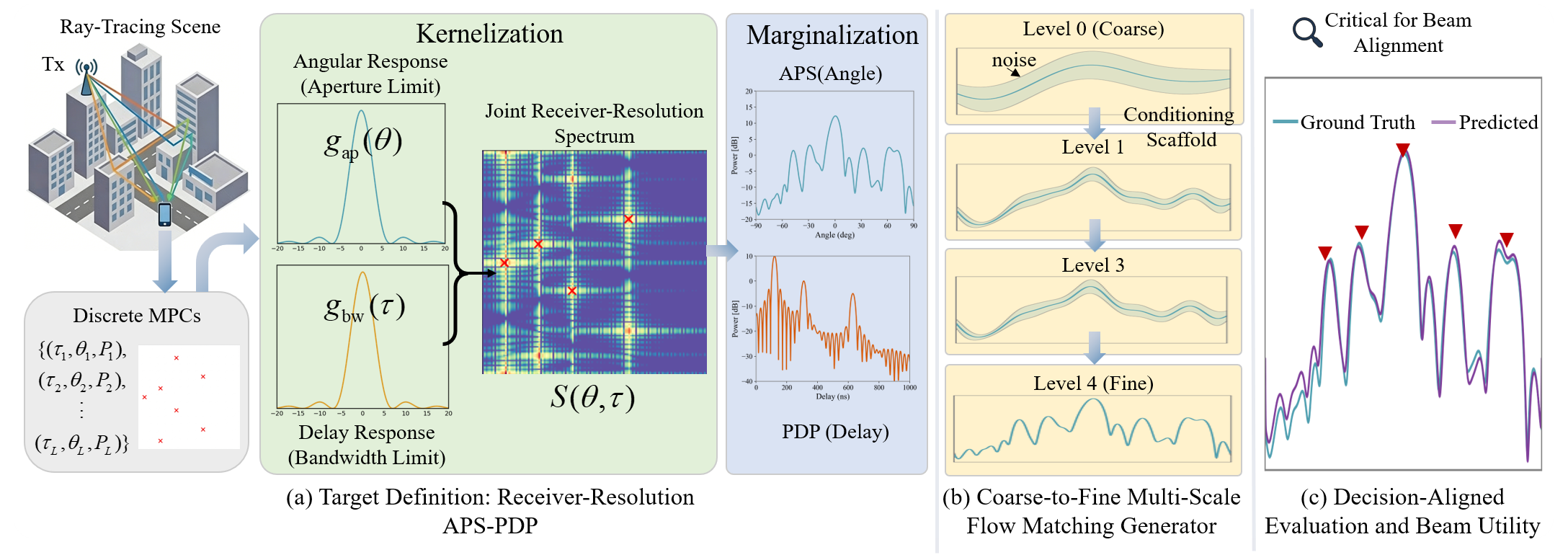}
  \captionof{figure}{\textbf{Receiver-resolution targets, coarse-to-fine generation, and decision-aligned evaluation.}
(a) Ray-traced multipath components are mapped to fixed-grid APS and PDP through explicit bandwidth and aperture kernels.
(b) \method generates spectra through a multi-scale flow-matching pyramid that refines dispersion envelopes into beam-relevant peaks and low-power tails.
(c) Evaluation connects spectral fidelity to beam utility and 3GPP moment errors.}
  \label{fig:teaser}
\end{minipage}

This issue becomes decisive when the goal shifts from coarse radio maps to spectrum maps. Coverage-oriented maps typically predict spatial averages such as received power, where moderate smoothing may have limited system impact \cite{Wang_2024_RadioDiff,Levie_2021,Ha_2024_RadioMapsBeamAlignment}. In contrast, spectrum maps must preserve the locations and sharpness of dominant peaks, as well as the distribution of weak sidelobes and long tails \cite{Ko_2017_PADP,3gpp_tr_38901}. Peak errors directly change beam ranking, while small mass shifts in weak tails can bias moment-based dispersion statistics \cite{Ha_2024_RadioMapsBeamAlignment,Ko_2017_PADP,3gpp_tr_38901}. The conditional spectrum can also be multimodal under the same available scene condition, since local geometry and unresolved structures may induce several plausible angle-delay patterns. Point estimators trained with squared error tend to average these modes and attenuate beam-relevant peaks \cite{Bishop_1994_MDN,Isola_2017_pix2pix}.

This reconstruction ambiguity stems from leaving receiver resolution implicit. The same MPC list can induce different spectra when bandwidth response, aperture response, or discretization changes. To make receiver-resolution learning and evaluation reproducible, we introduce \dataset, which fixes a receiver-domain mapping from ray-traced MPC lists to spectra. For each link, we apply a bandwidth-limited delay response and an aperture-limited angular response to kernelize the multipath set into a receiver-resolution angle-delay power spectrum, matching what a finite-bandwidth waveform and a finite-aperture array can observe \cite{Oppenheim_1996_SignalsSystems,VanTrees_2002_OAP}. We then marginalize this spectrum to obtain fixed-grid APS and PDP, and package them into a single tensor with an explicit validity mask. Thus, training losses and evaluation metrics operate on the same support. Fig.~\ref{fig:teaser}(a) summarizes this receiver-resolution target construction.

Built on \dataset, we propose \method, a coarse-to-fine multi-scale generator based on autoregressive flow matching. The model first samples a low-resolution spectrum that captures the dispersion envelope, then refines it across scales using conditional flow fields scaffolded by coarser outputs. This factorization separates coarse dispersion modeling from fine-scale refinement, while inference cost is measured by the number of function evaluations (NFE) for budget-controlled spectrum generation. Fig.~\ref{fig:teaser}(b) illustrates the coarse-to-fine refinement process.

Experiments on the test set show that \method improves decision-aligned beam utility and reduces errors of 3GPP dispersion statistics under a unified \(\mathrm{NFE}=80\) budget. Fig.~\ref{fig:teaser}(c) summarizes our decision-aligned evaluation, which connects spectral fidelity to beam ranking and 3GPP moment errors. Our contributions are summarized as follows:
\begin{enumerate}[label=(\roman*),leftmargin=2.55em,itemsep=0.08em,topsep=0.2em,parsep=0pt,partopsep=0pt]
\item We introduce \dataset, a receiver-resolution benchmark with explicit bandwidth and aperture kernels, fixed-grid APS/PDP targets, and a validity mask.
\item We propose \method, a coarse-to-fine autoregressive flow-matching generator that separates dispersion-envelope modeling from full-resolution peak and tail refinement.
\item We provide a decision-aligned evaluation protocol that connects spectral fidelity to beam ranking and 3GPP moment errors under a unified inference budget.
\end{enumerate}

\section{Related Work}
\label{sec:related}

\subsection{Receiver-Facing Spectra and 3GPP Metrics}
Receiver operation in wideband directional systems depends on how received energy is resolved over angle and delay under finite aperture and finite bandwidth \cite{Sayeed_2002,Brady_2013,Molisch2011Wireless,Akdeniz2014mmWave}. APS supports beam ranking and candidate selection \cite{Xiao_2016,Alkhateeb2014HybridPrecoding}, while PDP determines delay dispersion and equalization requirements \cite{Chiueh_2007}. Standard channel evaluation further summarizes these spectra through power-weighted moments such as delay spread (DS) and azimuth spread of arrival (ASA) \cite{3gpp_tr_38901,Jaeckel_2014_QuaDRiGa,Liu_2012_COST2100}. These links between receiver spectra, beam decisions, and dispersion moments motivate learning targets that are defined in the same observation domain as practical receivers. Our work follows this receiver-facing view and studies APS/PDP maps at the resolution induced by bandwidth and array aperture.

\subsection{Ray-Tracing Supervision and Radio Map Benchmarks}
Ray tracing provides scalable site-specific supervision for data-driven propagation modeling \cite{Hoydis_2023}. Existing channel datasets and simulators often release variable-cardinality multipath component lists, channel matrices, or link-level quantities \cite{Falade_2023,Alkhateeb_2019_DeepMIMO,Alrabeiah_2019_ViWi,Klautau2021Raymobtime,Jaeckel_2014_QuaDRiGa,Liu_2012_COST2100}. In parallel, radio map benchmarks have standardized scene splits and input-output protocols for coverage and pathloss prediction \cite{Levie_2021,Wang_2024_RadioDiff,Yapar_2024_PathlossChallenge,ChavesVillota_2023_DeepREM,Zhang_2023_RMEGAN,Sallouha_2024_REMUNet,Jia_2025_PhyRMDM,Wang_2025_RadioDiffk2}. These resources are valuable, but they usually do not fix a receiver-resolution APS/PDP target with explicit bandwidth and aperture kernels. As a result, spectrum learning and dispersion evaluation can still depend on post hoc reconstruction choices. \dataset addresses this gap by converting ray-traced supervision into fixed-grid APS/PDP maps with a shared receiver-domain target definition.

\subsection{Learning and Generative Modeling for Spectrum Synthesis}
Learning-based propagation surrogates have been developed for dense map querying and link-level inference under environmental conditioning \cite{Levie_2021,Liao_2026}. Single-shot regressors are efficient, but pointwise losses can blur spectra when multiple angular or delay patterns are consistent with the available condition. This becomes limiting when downstream decisions depend on peak location, relative peak strength, and low-power delay tails. Generative models provide a natural alternative for structured spectrum synthesis. Diffusion and score-based models have become standard iterative generators \cite{Ho2020DDPM,Song2020DDIM,Song2021ScoreSDE}, with Transformer-based denoisers further improving their scalability \cite{Peebles2023DiT}.
Flow matching learns continuous transport dynamics without simulating a forward diffusion process \cite{Lipman2023Flow,RectifiedFlow2022}. Recent autoregressive and coarse-to-fine formulations further separate global structure from fine residual detail \cite{ren2025flowar}. We adapt this idea to receiver-resolution APS/PDP maps, where coarse scales model dispersion envelopes and later scales refine decision-relevant spectral details.

\section{Method}
\label{sec:method}

\subsection{Overview}
\label{sec:method_overview}

The objective is to synthesize receiver-resolution APS/PDP maps that are directly usable for beam management and dispersion-aware wideband processing.
As shown in Fig.~\ref{fig:teaser}(a), we first map each variable-cardinality multipath component (MPC) list into fixed-grid APS and PDP targets, and then pack them into a unified two-channel tensor \(y\) with validity mask \(m\).
Given a condition tensor \(I\), the learning problem is to model the conditional distribution
\begin{align}
p_\theta(y\mid I),
\label{eq:cond_distribution}
\end{align}
where all losses and metrics are evaluated only on valid entries indicated by \(m\).
This fixed-shape representation provides a common interface for joint APS/PDP learning, batching, multi-scale modeling, and 3GPP-style evaluation.

\begin{figure}[t]
  \centering
  \includegraphics[width=\linewidth]{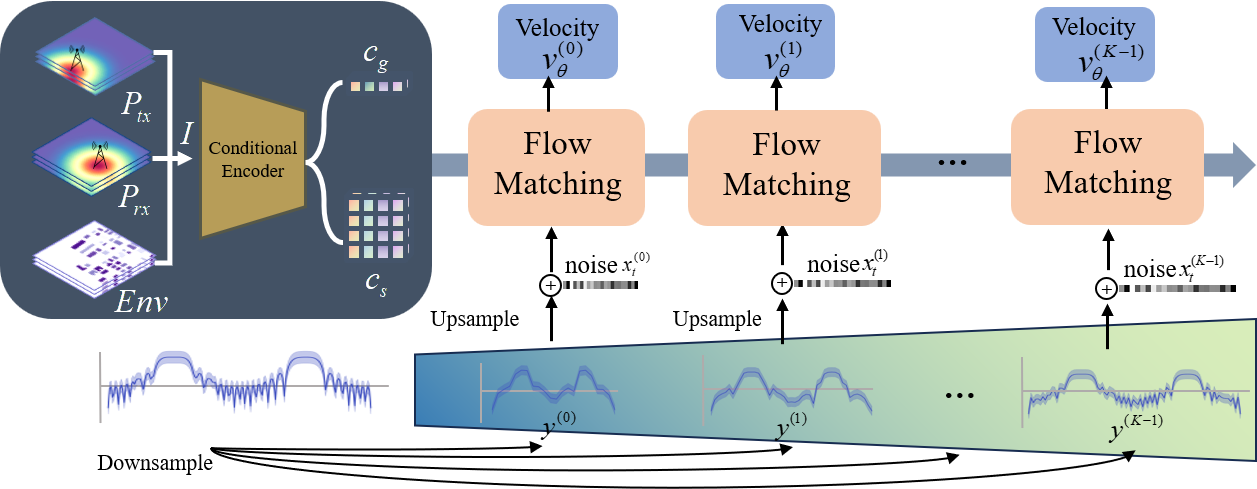}
  \caption{\textbf{Coarse-to-fine multi-scale generator in \method.}
The condition encoder transforms \(I\) into global context \(c_g\) and spatial tokens \(c_s\).
A \((K{+}1)\)-level autoregressive pyramid samples from coarse envelopes to full-resolution spectra via scale-wise conditional flow matching, where each scale is conditioned on the upsampled scaffold from the previous one.}
  \label{fig:model_arch}
\end{figure}

Built on this target definition, \method performs coarse-to-fine autoregressive flow matching.
As summarized in Fig.~\ref{fig:model_arch}, a condition encoder extracts global context and spatial tokens from \(I\).
A multi-scale pyramid then samples coarse dispersion envelopes first and progressively refines them into full-resolution spectral details.
At each scale, a conditional velocity field transports Gaussian noise toward the target spectrum under the current condition and the previous-scale scaffold.
The full velocity-network architecture is provided in Appendix~\ref{app:velnet}.

\subsection{Receiver-Resolution Supervision from MPCs}
\label{sec:method_targets}

Each link is associated with an unordered multipath set
\(\mathcal{P}=\{(a_\ell,\tau_\ell,\phi_\ell)\}_{\ell=1}^{L}\),
where \(a_\ell\in\mathbb{C}\) is the complex gain, \(\tau_\ell\) is the geometric delay, and \(\phi_\ell\) is the angle of arrival.
Ray tracing typically outputs \(\mathcal{P}\) as a variable-cardinality list, while a practical receiver observes energy only after finite-bandwidth filtering and finite-aperture beamforming.
We therefore define supervision directly in the receiver observation domain.

We kernelize each MPC into the discretized angle-delay plane and sum power contributions:
\begin{align}
S(\theta,\tau)
=
\sum_{\ell=1}^{L}
|a_\ell|^2
\bigl|g_{\mathrm{bw}}(\tau-\tau_\ell)\bigr|^2
\bigl|g_{\mathrm{ap}}(\theta;\phi_\ell)\bigr|^2 ,
\label{eq:adps}
\end{align}
where \(g_{\mathrm{bw}}\) models the bandwidth-limited delay response and \(g_{\mathrm{ap}}\) models the aperture-limited angular response.
This power-domain target is determined by the assumed receiver bandwidth and array aperture, rather than by post hoc reconstruction choices.

We use the sinc delay kernel and the normalized ULA response:
\begin{align}
g_{\mathrm{bw}}(\Delta\tau)
=
\mathrm{sinc}(B\Delta\tau),
\qquad
g_{\mathrm{ap}}(\theta;\phi)
=
\frac{\sin(N\Psi)}{N\sin\Psi},
\label{eq:kernels}
\end{align}
where \(B\) is the system bandwidth, \(N\) is the number of ULA elements, and
\(\Psi(\theta,\phi)=\pi(d/\lambda)(\sin\theta-\sin\phi)\), with element spacing \(d\) and wavelength \(\lambda\).
At \(\Psi=0\), the continuous limit of \(g_{\mathrm{ap}}\) is used.

From \(S(\theta,\tau)\), we obtain receiver-facing spectra by marginalization:
\begin{align}
A(\theta)
=
\sum_{\tau} S(\theta,\tau),
\qquad
D(\tau)
=
\sum_{\theta} S(\theta,\tau).
\label{eq:aps_pdp}
\end{align}
Here \(A(\theta)\) is the APS and \(D(\tau)\) is the PDP on fixed receiver grids.
Both spectra are derived from the same angle-delay power spectrum, which keeps angular and delay statistics aligned.

For joint modeling, we pack APS and PDP into a unified two-channel target
\(y\in\mathbb{R}^{N_\theta\times 2}\):
\begin{align}
y[:,0]
=
A,
\qquad
y[:,1]
=
\Pi(D),
\label{eq:joint_pack}
\end{align}
where \(\Pi:\mathbb{R}^{N_\tau}\rightarrow\mathbb{R}^{N_\theta}\) embeds the PDP bins into a fixed support on the length-\(N_\theta\) axis.
Entries outside the embedded PDP support are padded with zeros.
The corresponding validity mask is
\begin{align}
m[:,0]
\equiv
1,
\qquad
m[:,1]
=
\Pi(\mathbf{1}),
\label{eq:joint_mask}
\end{align}
where \(\mathbf{1}\in\mathbb{R}^{N_\tau}\) is the all-ones vector.
The mask excludes padded PDP entries from losses and metrics.

\subsection{Coarse-to-Fine Autoregressive Generator}
\label{sec:method_arch}

The model consists of a condition encoder and a multi-scale autoregressive generator operating on the packed joint spectrum \(y\).
The encoder maps \(I\) to a global context vector \(c_g\) and spatial tokens \(c_s\).
The global context captures scene-level information shared by all scales, while spatial tokens preserve localized geometric cues.

We construct a \((K{+}1)\)-level pyramid \(\{y^{(k)}\}_{k=0}^{K}\) by downsampling the packed spectrum along its length axis while keeping the two-channel layout.
Scale \(k=0\) represents coarse dispersion envelopes, and scale \(k=K\) represents full-resolution peak-and-tail structure.
The conditional distribution is factorized as
\begin{align}
p_\theta(y\mid I)
=
p_\theta\!\bigl(y^{(0)}\mid I\bigr)
\prod_{k=1}^{K}
p_\theta\!\bigl(y^{(k)}\mid y^{(k-1)}, I\bigr).
\label{eq:ms_factor}
\end{align}
This factorization turns joint APS/PDP generation into progressive refinement on a fixed receiver-resolution interface.

At each scale \(k\), generation is defined by a conditional ODE:
\begin{align}
\dot{x}^{(k)}(t)
=
v_\theta^{(k)}
\!\left(
x^{(k)}(t),t;I,y^{(k-1)}
\right),
\qquad
t\in[0,1],
\label{eq:scale_ode}
\end{align}
where \(v_\theta^{(k)}\) is the scale-wise velocity field.
For \(k=0\), the previous-scale condition is absent.
For \(k\ge 1\), the model conditions on an upsampled scaffold from \(y^{(k-1)}\).
The velocity field is implemented by an AdaLN Transformer that processes noisy spectrum tokens with the cross-scale scaffold and condition features.
Further architectural details are provided in Appendix~\ref{app:velnet}.

\subsection{Multi-Scale Flow Matching Objective}
\label{sec:method_fm}

We train the scale-wise velocity fields with flow matching.
At scale \(k\), we draw a target spectrum \(y^{(k)}\), a Gaussian source sample
\(x_0^{(k)}\sim\mathcal{N}(0,\mathbf{I})\), and a time \(t\sim\mathcal{U}[0,1]\).
We use the linear probability path
\begin{align}
x_t^{(k)}
=
(1-t)x_0^{(k)}
+
t y^{(k)},
\label{eq:prob_path}
\end{align}
whose target velocity is \(u_t^{(k)}=y^{(k)}-x_0^{(k)}\).

Let \(m^{(k)}\) be the validity mask at scale \(k\), obtained by applying the same scale transformation to the full-resolution mask.
The training loss is
\begin{align}
\mathcal{L}
=
\sum_{k=0}^{K}
\alpha_k
\,
\mathbb{E}
\Bigl[
\bigl\|
\bigl(
v_\theta^{(k)}
(
x_t^{(k)},t;I,y^{(k-1)}
)
-
u_t^{(k)}
\bigr)
\odot m^{(k)}
\bigr\|^2
\Bigr],
\label{eq:ms_fm_loss}
\end{align}
with the convention that \(y^{(k-1)}\) is absent at \(k=0\).
This objective aligns the learned dynamics with the valid receiver-resolution support at every scale.

\subsection{Coarse-to-Fine Sampling and Compute Control}
\label{sec:method_sampling}

At inference, we first sample \(y^{(0)}\) by integrating the scale-\(0\) ODE from \(t=0\) to \(t=1\), starting from Gaussian noise.
For \(k\ge 1\), we upsample the previous-scale sample as a scaffold and integrate the scale-\(k\) ODE conditioned on this scaffold.
If scale \(k\) uses \(S_k\) solver steps, the total sampling budget is
\begin{align}
\mathrm{NFE}
=
\sum_{k=0}^{K} S_k .
\label{eq:nfe}
\end{align}
This definition gives an explicit sampling-budget control for latency-quality tradeoffs and is used for all budget-matched comparisons.

\section{Experiments}
\label{sec:experiments}

We evaluate receiver-resolution APS/PDP map generation on \dataset.
All methods use the same receiver-resolution target \(y\in\mathbb{R}^{N_\theta\times 2}\) and the same validity mask \(m\).
The main quantitative comparison uses the test split with a unified metric pipeline.
This protocol keeps the target construction, valid support, and evaluation metrics consistent across deterministic baselines, neural regressors, diffusion models, and flow-matching models.
We report beam utility, 3GPP dispersion errors, spectral fidelity, and inference latency under a unified receiver-resolution evaluation protocol.
The main iterative comparison uses a unified \(\mathrm{NFE}=80\) budget.
We further evaluate iterative generators under multiple sampling budgets to study latency-quality behavior.
Implementation details are provided in Appendix~\ref{app:impl}, and supplementary results are reported in the appendix.

\subsection{Dataset and Receiver-Resolution Targets}
\label{sec:exp_dataset}

Each example in \dataset contains a condition tensor \(I\), a packed receiver-resolution target \(y\), and a binary validity mask \(m\).
The condition tensor encodes the environment and the queried Tx-Rx pair.
The first target channel stores the angle power spectrum (APS) on the full angular grid, while the second target channel stores the power delay profile (PDP) embedded into a fixed support on the same length axis.
Entries outside the embedded PDP support are padded with zeros and excluded by the validity mask.
All models are trained and evaluated with the same packing rule and validity mask.

\begin{figure}[t]
  \centering
  \includegraphics[width=0.98\linewidth]{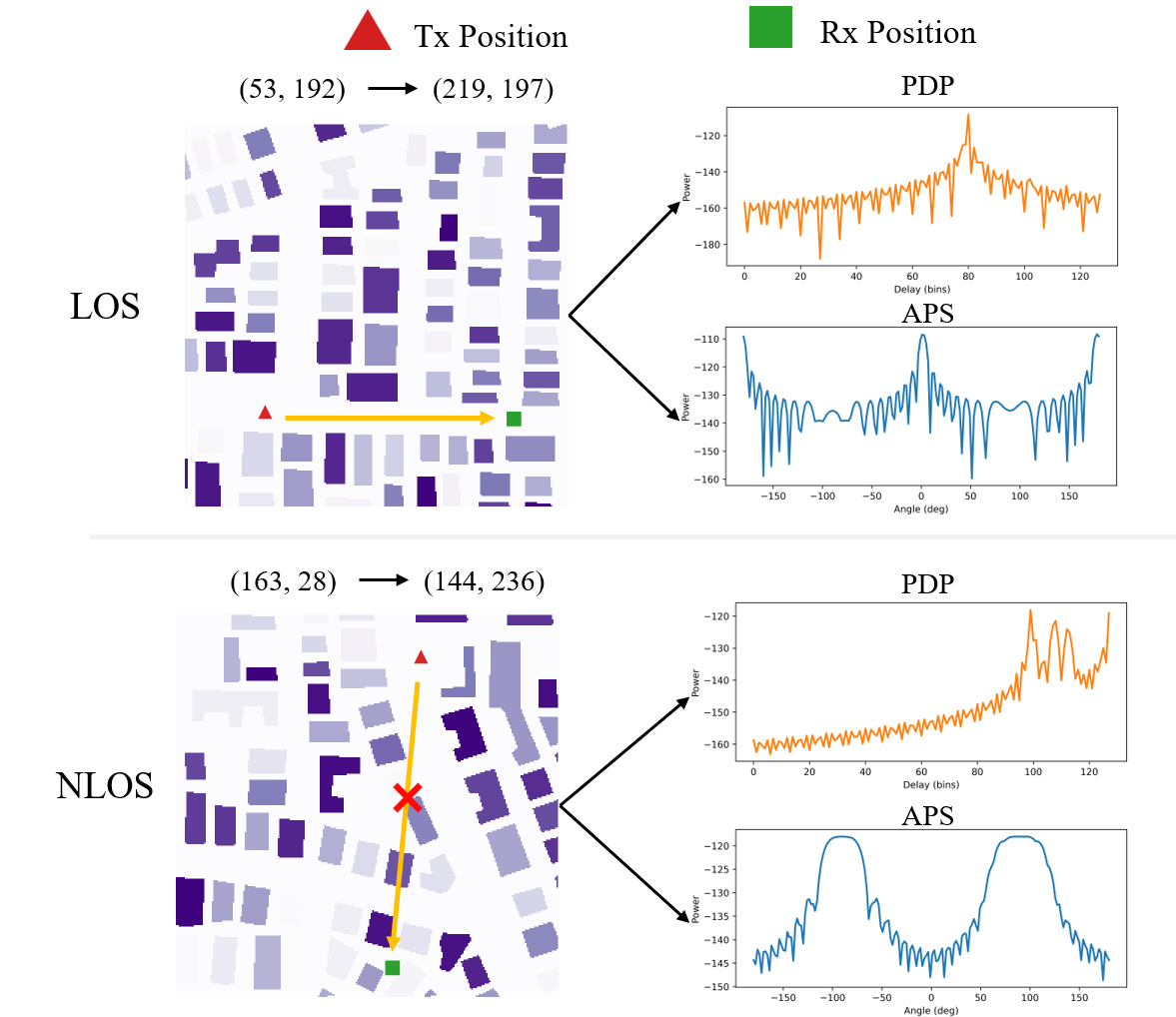}
  \caption{\textbf{Representative receiver-resolution targets in \dataset.}
  Left: condition tensor \(I\) with Tx and queried Rx on the occupancy map.
  Middle: receiver-resolution PDP.
  Right: receiver-resolution APS.
  LoS and NLoS links differ in peak location, angular spread, delay onset, and low-power tail structure.}
  \label{fig:data_example}
  \vspace{-2.3em}
\end{figure}

Unless stated otherwise, the receiver configuration follows the target construction in Sec.~\ref{sec:method_targets}.
We use carrier frequency \(f_c=2.4~\mathrm{GHz}\), bandwidth \(B=200~\mathrm{MHz}\), and a 16-element uniform linear array (ULA) with half-wavelength spacing.
The angle-delay power spectrum is discretized into \(N_\tau\times N_\theta=128\times180\) bins, with delay range \([0,1000]~\mathrm{ns}\) and angular range \([-180^\circ,180^\circ)\).
These settings define the receiver-resolution APS/PDP targets used by all methods.
\dataset contains \(S=50\) urban scenes, with \(50{,}000\) Tx-Rx links per scene and \(2.5\) million links in total.
We partition the scenes into 40 for training, 5 for validation, and 5 for testing.
All links from the same scene are assigned to the same split, so test examples come from urban layouts not used during training.
Figure~\ref{fig:data_example} shows representative receiver-resolution targets from \dataset.
The examples include the condition tensor with the Tx-Rx query, the receiver-resolution PDP, and the receiver-resolution APS.
They illustrate that links with different propagation conditions can differ in dominant peak location, angular spread, delay onset, and low-power tail structure.
These structures are not captured by average received power maps, but they directly affect beam ranking and dispersion statistics.
The benchmark therefore evaluates both spectral fidelity and receiver-decision metrics derived from APS and PDP.

\subsection{Baselines}
\label{sec:exp_baselines}

We compare \method with five baselines.
RT-LoS+Edge is a deterministic sparse-geometry reference.
ChannelGAN and UNet are single-shot neural baselines trained on the same packed APS/PDP target.
DiT is a diffusion Transformer baseline~\cite{Peebles2023DiT} evaluated under the same NFE budget.
Single-scale FM keeps the same flow-matching objective and receiver-spectrum interface as \method, but removes coarse-to-fine autoregressive refinement.
It therefore serves as the main architecture ablation.
Detailed architectures and training settings are given in Appendix~\ref{app:baseline_details}.

\subsection{Metrics}
\label{sec:exp_metrics}

We evaluate three aspects of receiver-resolution APS/PDP maps.
First, beam utility is measured by BPL@5, which evaluates the power loss of the top beam candidates selected from the predicted APS.
Second, 3GPP-style dispersion accuracy is measured by absolute errors of delay spread (DS) and azimuth spread of arrival (ASA), computed from the predicted PDP and APS.
Third, spectral fidelity is measured by APS RMSE, APS JSD, PDP RMSE, and PDP JSD.
All metrics are computed in the linear power domain after applying the validity mask.
Metric formulas are provided in Appendix~\ref{app:metrics}.

\subsection{Inference Budget and Timing}
\label{sec:exp_budget}

We measure sampling compute using the number of function evaluations (NFE).
For \method, the scale \(k\) ODE uses \(S_k\) solver steps, and the total budget is
\begin{align}
\mathrm{NFE}
=
\sum_{k=0}^{K} S_k .
\label{eq:nfe_impl}
\end{align}
The default setting uses four scales with 20 solver steps at each scale, giving \(\mathrm{NFE}=80\).
For Single-scale FM and DiT, NFE equals the number of sampling steps.
The main comparison uses a unified \(\mathrm{NFE}=80\) budget.
We also sweep \(\mathrm{NFE}\in\{12,20,40,60,80\}\) for the iterative generators to evaluate sampling-budget sensitivity.
Latency is measured as per-sample sampling time with GPU synchronization and excludes disk I/O and host-device transfer.
The detailed timing protocol is provided in Appendix~\ref{app:latency}.

\vspace{-1em}
\section{Results}
\label{sec:results}

\subsection{Decision-Aligned APS/PDP Map Generation}
\label{sec:res_main}

Table~\ref{tab:main_decision_disp} reports the main test-set comparison under the unified \(\mathrm{NFE}=80\) budget, using the three metrics most directly tied to receiver decisions and 3GPP dispersion: BPL@5, \(|\Delta\mathrm{DS}|\), and \(|\Delta\mathrm{ASA}|\).

\begin{table}[t]
\caption{\textbf{Test-set beam utility and 3GPP dispersion errors.}
Lower values are better. BPL@5 is measured in dB. DS and ASA are absolute errors.}
\label{tab:main_decision_disp}
\centering
\begingroup
\setlength{\tabcolsep}{4pt}
\renewcommand{\arraystretch}{1.08}
\begin{tabular*}{\linewidth}{@{\extracolsep{\fill}}lccc@{}}
\toprule
Method & BPL@5 \(\downarrow\) & \(|\Delta\mathrm{DS}|\) (ns) \(\downarrow\) & \(|\Delta\mathrm{ASA}|\) (deg) \(\downarrow\) \\
\midrule
RT-LoS+Edge & 6.32 & 56.95 & 71.13 \\
ChannelGAN & 21.03 & 278.61 & 85.18 \\
UNet & 13.92 & 299.93 & 148.66 \\
DiT & 6.23 & 33.43 & 20.67 \\
Single-scale FM & 2.42 & 29.76 & 11.87 \\
\midrule
\method & \textbf{2.03} & \textbf{29.12} & \textbf{10.82} \\
\bottomrule
\end{tabular*}
\endgroup
\end{table}

\method achieves the lowest BPL@5, \(|\Delta\mathrm{DS}|\), and \(|\Delta\mathrm{ASA}|\) among the compared methods.
Single-scale FM is the closest learned baseline, which indicates that flow matching is already a strong generator for receiver-resolution spectra.
The additional gain of \method is most visible on beam utility and angular dispersion, rather than on every pointwise fidelity metric.
RT-LoS+Edge provides a sparse-geometry reference, but its large ASA error shows that a small set of selected paths is insufficient to recover receiver-facing angular statistics.
Scene-level confidence intervals are reported in Appendix~\ref{app:ci}.

\subsection{Architecture Ablation}
\label{sec:res_ablation}

Single-scale FM is the main architecture ablation of \method.
It keeps the same receiver-resolution target and masked flow-matching objective, but removes the coarse-to-fine autoregressive pyramid.
This makes the comparison isolate the effect of progressive refinement under the same target definition.
As shown in Table~\ref{tab:main_decision_disp}, \method improves BPL@5, \(|\Delta\mathrm{DS}|\), and \(|\Delta\mathrm{ASA}|\) over Single-scale FM under the unified \(\mathrm{NFE}=80\) budget.
Table~\ref{tab:main_fidelity} shows a complementary pattern: Single-scale FM has lower APS RMSE, while \method gives lower APS JSD, PDP RMSE, and PDP JSD.
This contrast supports the receiver-facing evaluation design, since pointwise error alone does not fully capture beam ranking or dispersion statistics.

\subsection{Sampling Budget Analysis}
\label{sec:res_nfe_sweep}

Table~\ref{tab:nfe_sweep_values} compares iterative generators under matched NFE values.
The main text reports the same decision and dispersion metrics used in Table~\ref{tab:main_decision_disp}.
The full NFE sweep with APS/PDP RMSE and JSD is provided in Appendix~\ref{app:nfe_sweep_full}.

\begin{table*}[t]
\caption{\textbf{Sampling-budget analysis under matched NFE values.}
Each block fixes the sampling budget and compares iterative generators under the same metrics.
Lower values are better.}
\label{tab:nfe_sweep_values}
\centering
\begingroup
\setlength{\tabcolsep}{7pt}
\renewcommand{\arraystretch}{1.08}
\begin{tabular*}{\textwidth}{@{\extracolsep{\fill}}clccc@{}}
\toprule
NFE & Method & APS BPL@5 \(\downarrow\) & APS \(|\Delta\mathrm{ASA}|\) \(\downarrow\) & PDP \(|\Delta\mathrm{DS}|\) \(\downarrow\) \\
\midrule
\multirow{3}{*}{12}
& DiT & 4.92 & 21.95 & 44.07 \\
& Single-scale FM & \textbf{1.74} & 10.90 & 31.42 \\
& \method & 1.81 & \textbf{10.89} & \textbf{28.61} \\
\addlinespace[2pt]
\midrule

\multirow{3}{*}{20}
& DiT & 5.01 & 18.80 & 36.24 \\
& Single-scale FM & 1.96 & 11.07 & 30.11 \\
& \method & \textbf{1.89} & \textbf{10.65} & \textbf{29.12} \\
\addlinespace[2pt]
\midrule

\multirow{3}{*}{40}
& DiT & 5.46 & 18.99 & 33.66 \\
& Single-scale FM & 2.18 & 11.43 & 29.76 \\
& \method & \textbf{1.96} & \textbf{10.59} & \textbf{29.48} \\
\addlinespace[2pt]
\midrule

\multirow{3}{*}{60}
& DiT & 5.60 & 19.19 & 33.55 \\
& Single-scale FM & 2.27 & 11.28 & 29.75 \\
& \method & \textbf{2.04} & \textbf{10.98} & \textbf{29.49} \\
\addlinespace[2pt]
\midrule

\multirow{3}{*}{80}
& DiT & 6.23 & 20.67 & 33.43 \\
& Single-scale FM & 2.42 & 11.87 & 29.76 \\
& \method & \textbf{2.03} & \textbf{10.82} & \textbf{29.12} \\
\bottomrule
\end{tabular*}
\endgroup
\vspace{-1.5em}
\end{table*}

At \(\mathrm{NFE}=12\), Single-scale FM gives the lowest APS BPL@5, while \method gives the lowest APS \(|\Delta\mathrm{ASA}|\) and PDP \(|\Delta\mathrm{DS}|\).
At \(\mathrm{NFE}\in\{20,40,60,80\}\), \method gives the lowest reported values on all three main metrics.
The sweep also shows that increasing NFE does not yield a universal monotonic gain.
The preferred operating point depends on the metric, so NFE should be treated as a matched sampling budget rather than a per-method tuning advantage.

\subsection{Spectral Fidelity}
\label{sec:res_fidelity}

Table~\ref{tab:main_fidelity} reports spectral fidelity on APS and PDP.
This table complements the decision-aligned comparison in Table~\ref{tab:main_decision_disp}.

\begin{table}[t]
\caption{\textbf{Spectral fidelity on APS and PDP.}
Lower values are better. JSD denotes Jensen-Shannon divergence.}
\label{tab:main_fidelity}
\centering
\begingroup
\setlength{\tabcolsep}{4pt}
\renewcommand{\arraystretch}{1.08}
\begin{tabular*}{\linewidth}{@{\extracolsep{\fill}}lcccc@{}}
\toprule
& \multicolumn{2}{c}{APS \(\downarrow\)} & \multicolumn{2}{c}{PDP \(\downarrow\)} \\
\cmidrule(lr){2-3}\cmidrule(lr){4-5}
Method & RMSE & JSD & RMSE & JSD \\
\midrule
RT-LoS+Edge & 0.082 & 0.174 & \textbf{0.037} & 0.243 \\
ChannelGAN & 0.096 & 0.681 & 0.089 & 0.692 \\
UNet & 0.338 & 0.095 & 0.090 & 0.639 \\
DiT & 0.043 & 0.199 & 0.093 & 0.229 \\
Single-scale FM & \textbf{0.021} & 0.082 & 0.079 & 0.195 \\
\midrule
\method & 0.050 & \textbf{0.073} & 0.078 & \textbf{0.187} \\
\bottomrule
\end{tabular*}
\endgroup
\end{table}

The fidelity metrics reveal a different ranking from the decision-aligned metrics.

Single-scale FM obtains the lowest APS RMSE among the iterative generators, while \method obtains the lowest APS JSD and PDP JSD, and also gives the lowest PDP RMSE among the iterative generators.
This contrast shows that pointwise error and receiver-facing metrics can favor different spectral properties.

RT-LoS+Edge has the lowest PDP RMSE overall, but its BPL@5 and ASA error in Table~\ref{tab:main_decision_disp} show that low pointwise delay error alone does not imply reliable receiver-facing decisions.

Thus, Table~\ref{tab:main_fidelity} complements rather than replaces the decision and dispersion comparison in Table~\ref{tab:main_decision_disp}.
Qualitative APS comparisons are provided in Appendix~\ref{app:qualitative}.

\subsection{Efficiency and Supplementary Analyses}
\label{sec:res_supp}

\begin{wrapfigure}[14]{r}{0.50\linewidth}
  \vspace{-3.0em}
  \centering
  \captionsetup{font=small,skip=2pt}
  \includegraphics[width=\linewidth]{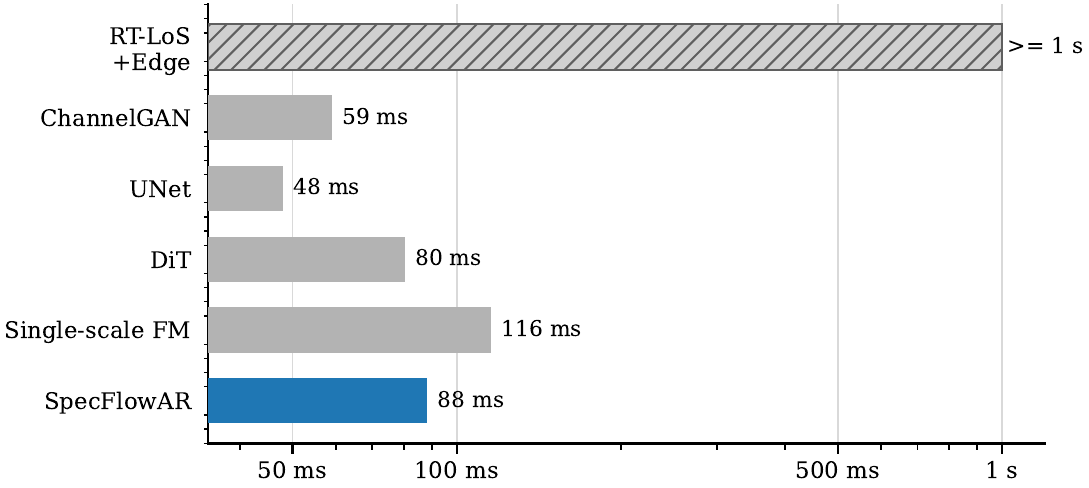}
  \caption{\textbf{APS inference latency.}
  Runtime is measured per sample with GPU synchronization, excluding disk I/O and host-device transfer.
  Iterative generators use the unified \(\mathrm{NFE}=80\) budget, while ChannelGAN and UNet use one forward pass.
  RT-LoS+Edge is clipped at \(1~\mathrm{s}\) for visualization.
  The horizontal axis uses a logarithmic scale.}
  \label{fig:latency}
  \vspace{-0.5em}
\end{wrapfigure}

Figure~\ref{fig:latency} reports APS inference latency under the same timing protocol used in Sec.~\ref{sec:exp_budget}.
The latency result should be read with the receiver-facing metrics in Table~\ref{tab:main_decision_disp}.
DiT is slightly faster than \method, but it has higher BPL@5, \(|\Delta\mathrm{DS}|\), and \(|\Delta\mathrm{ASA}|\).
Compared with Single-scale FM, \method is faster and also improves the main decision and dispersion metrics.
RT-LoS+Edge has substantially higher latency, which limits its use for dense spectrum queries.
These results show that \method improves receiver-resolution APS/PDP generation while keeping APS inference practical under the matched evaluation protocol.

\section{Conclusion}
\label{sec:conclusion}

We have introduced \dataset, a receiver-resolution benchmark that maps ray-traced MPC lists to fixed-grid APS/PDP targets using fixed bandwidth and aperture kernels with a validity mask.
We have further developed \method, a coarse-to-fine autoregressive flow-matching generator that refines beam-relevant peaks and low-power tails under explicit sampling-budget control.
Experiments have shown improved beam utility and more accurate 3GPP dispersion moments under matched inference budgets, supporting dense spectrum queries in urban wireless environments.
The current protocol fixes the receiver kernels, bandwidth, aperture, and grids for controlled comparison, while future work will calibrate these choices with measured channels and extend them to variable hardware settings.

\newpage
\bibliographystyle{plainnat}
\bibliography{example_paper}

\clearpage
\appendix
\numberwithin{figure}{section}
\numberwithin{table}{section}

\numberwithin{equation}{section}
\renewcommand{\theequation}{\thesection.\arabic{equation}}
\renewcommand{\thefigure}{\thesection.\arabic{figure}}
\renewcommand{\thetable}{\thesection.\arabic{table}}

\section{Evaluation Metrics}
\label{app:metrics}

This section defines the metrics used in the main text and supplementary tables.
For a prediction \(\widehat{y}\), we recover \(\widehat{A}\) and \(\widehat{D}\) from the packed representation and compare them with the target spectra \(A\) and \(D\).

\paragraph{Beam power loss.}
Beam utility is measured by beam power loss at top \(K\), denoted as BPL@\(K\).
Given the predicted APS \(\widehat{A}\), we rank the receive codebook \(\{\theta_b\}_{b=1}^{N_b}\) by \(\widehat{A}(\theta_b)\), and select the top \(K\) beams as \(\widehat{\mathcal{B}}_K\).
The oracle beam is selected by the target APS \(A\).
BPL@\(K\) is defined as
\begin{align}
\mathrm{BPL@}K
=
10\log_{10}
\frac{
\max_b A(\theta_b)
}{
\max_{b\in\widehat{\mathcal{B}}_K} A(\theta_b)
}.
\label{eq:app_bpl}
\end{align}
A smaller BPL@\(K\) indicates that the predicted top-\(K\) set contains a beam with ground-truth power closer to the oracle beam.

\paragraph{Delay spread.}
Delay spread (DS) is computed from the PDP as a power-weighted moment, following the standard RMS delay spread definition~\cite{3gpp_tr_38901}.
Let \(D_i\) be the PDP power at delay bin \(\tau_i\), and let
\begin{align}
w_i
=
\frac{D_i}{\sum_i D_i}.
\label{eq:app_delay_weight}
\end{align}
The delay spread is
\begin{align}
\mathrm{DS}(D)
=
\sqrt{
\sum_i w_i \tau_i^2
-
\left(
\sum_i w_i \tau_i
\right)^2
}.
\label{eq:app_ds}
\end{align}
We report the absolute DS error
\begin{align}
|\Delta\mathrm{DS}|
=
\left|
\mathrm{DS}(\widehat{D})
-
\mathrm{DS}(D)
\right|.
\label{eq:app_ds_error}
\end{align}

\paragraph{Azimuth spread of arrival.}
Azimuth spread of arrival (ASA) is computed from the APS using the circular standard deviation adopted for angular spread in 3GPP channel evaluation~\cite{3gpp_tr_38901}.
Let \(A_j\) be the APS power at azimuth bin \(\theta_j\), and let
\begin{align}
v_j
=
\frac{A_j}{\sum_j A_j}.
\label{eq:app_angle_weight}
\end{align}
We compute the first circular moment
\begin{align}
z
=
\sum_j v_j \exp(\mathrm{i}\theta_j),
\label{eq:app_circular_moment}
\end{align}
and define the angular spread as
\begin{align}
\mathrm{ASA}(A)
=
\sqrt{-2\log |z|}.
\label{eq:app_asa}
\end{align}
Angles are converted to degrees when reporting \(|\Delta\mathrm{ASA}|\).
The absolute ASA error is
\begin{align}
|\Delta\mathrm{ASA}|
=
\left|
\mathrm{ASA}(\widehat{A})
-
\mathrm{ASA}(A)
\right|.
\label{eq:app_asa_error}
\end{align}

\paragraph{RMSE.}
For a spectrum \(x\) and prediction \(\widehat{x}\), RMSE is computed over the evaluated bins as
\begin{align}
\mathrm{RMSE}
=
\sqrt{
\frac{1}{N}
\sum_{i=1}^{N}
\left(
\widehat{x}_i-x_i
\right)^2
}.
\label{eq:app_rmse}
\end{align}
We report APS RMSE and PDP RMSE separately.

\paragraph{Jensen-Shannon divergence.}
Jensen-Shannon divergence (JSD) measures the distributional mismatch between predicted and target spectra~\cite{Lin1991JSD,Men_ndez_1997}.
For two discrete spectral distributions \(p\) and \(q\), let
\begin{align}
r
=
\frac{1}{2}(p+q).
\label{eq:app_jsd_mixture}
\end{align}
The JSD is
\begin{align}
\mathrm{JSD}(p,q)
=
\frac{1}{2}
\mathrm{KL}(p\|r)
+
\frac{1}{2}
\mathrm{KL}(q\|r).
\label{eq:app_jsd}
\end{align}
We report APS JSD and PDP JSD separately.

\section{Velocity Network Architecture}
\label{app:velnet}
\begin{figure}[t]
  \centering
  \includegraphics[width=0.98\linewidth]{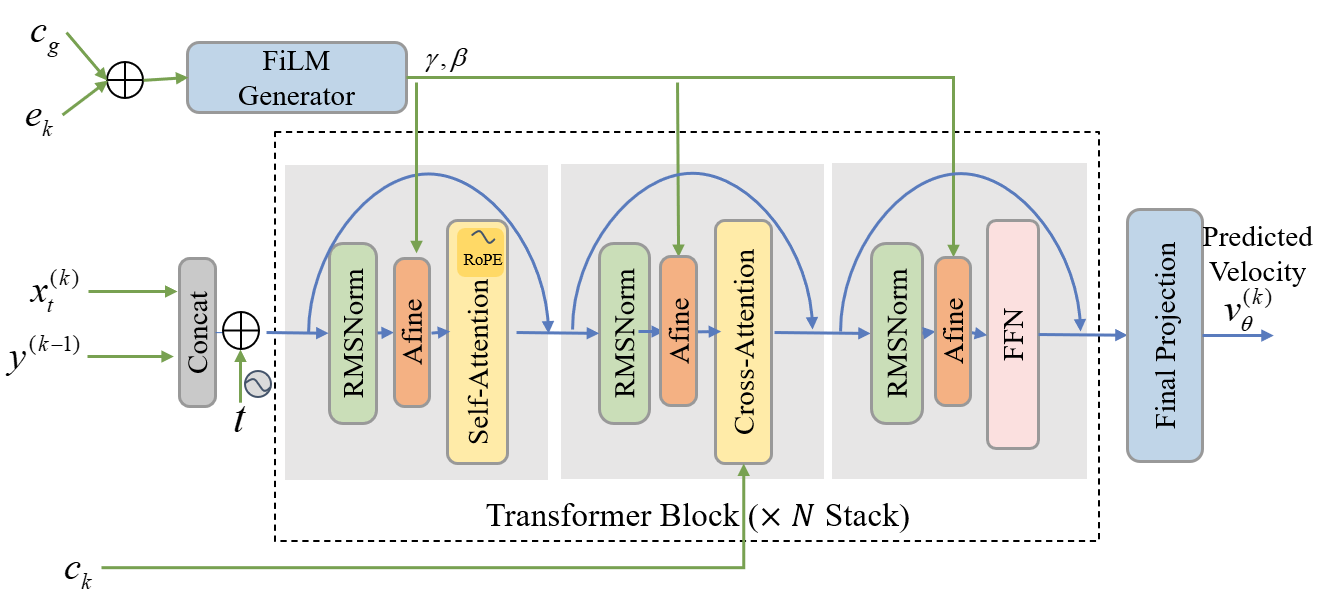}
  \caption{\textbf{Velocity Transformer for scale-wise flow matching.}
Noisy spectrum tokens are fused with the cross-scale scaffold and processed by an AdaLN Transformer~\cite{Peebles2023DiT}.
Global context modulates intermediate features, while cross-attention to spatial tokens injects localized geometry for peak and tail refinement.}
  \label{fig:vel_net}
\end{figure}

Figure~\ref{fig:vel_net} shows the scale-wise velocity network used in \method.
At scale \(k\), the input consists of noisy spectrum tokens \(x_t^{(k)}\), the time embedding of \(t\), the condition features from \(I\), and the upsampled scaffold from the previous scale when \(k\ge 1\).
The scaffold is concatenated with the noisy spectrum tokens along the feature dimension, so that the velocity field predicts residual transport relative to the coarse spectrum structure.
The global context \(c_g\) modulates Transformer blocks through AdaLN-style affine parameters.
The spatial tokens \(c_s\) are injected through cross-attention and provide localized geometric cues for peak and tail refinement.
Self-attention is applied along the packed spectrum axis, which allows APS bins and embedded PDP bins to exchange information within the same fixed-shape representation.
The network outputs a velocity tensor with the same shape as the scale-\(k\) packed spectrum, and the validity mask is applied only in the loss.

\section{Implementation Details}
\label{app:impl}

This section provides implementation details for \method.
All models are trained and evaluated on the same packed receiver-resolution target and validity mask described in Sec.~\ref{sec:method_targets}.

\paragraph{Condition tensor.}
The condition input is \(I\in\mathbb{R}^{3\times256\times256}\).
It stacks one occupancy channel and two endpoint channels that encode the transmitter and queried receiver locations.
The endpoint channels are rendered as Gaussian heatmaps and clipped to \([0,1]\).
This representation provides spatial context and link endpoints in a fixed image-like format.

\paragraph{Condition encoder.}
The condition encoder is a strided convolutional network.
It downsamples \(I\) to a \(16\times16\) feature map and produces a global context vector and \(M=256\) spatial tokens.
The hidden dimension is \(D=768\).
The global context is used for scene-level conditioning, while spatial tokens are used by the velocity Transformer through cross-attention.

\paragraph{Spectrum pyramid.}
We use a four-scale pyramid for coarse-to-fine generation.
The APS angular lengths are \(\{22,45,90,180\}\), and the PDP delay lengths are \(\{16,32,64,128\}\).
At each scale, the APS and PDP are packed into the same two-channel representation as the full-resolution target.
The scale-wise spectra are constructed by one-dimensional interpolation on their native axes.
Masks are resized with nearest-neighbor interpolation and packed with the same rule as the spectra.

\paragraph{Velocity Transformer.}
The velocity Transformer has 12 blocks, width 768, and 12 attention heads.
Its structural details are given in Sec.~\ref{app:velnet}.

\paragraph{Training.}
All \method models are trained with the masked multi-scale flow-matching objective in Eq.~\eqref{eq:ms_fm_loss}.
We use AdamW with learning rate \(10^{-4}\), weight decay \(10^{-2}\), and 200 training epochs.
Automatic mixed precision is used during training.

\paragraph{Compute resources.}
All reported training runs were conducted on a single NVIDIA H100 GPU with 80 GB HBM3 memory.
The host machine used an Intel Xeon Platinum 8562Y+ CPU with 1.0 TiB RAM.
The implementation used PyTorch 2.10.0+cu128, with a PyTorch CUDA build based on CUDA 12.8 and local \texttt{nvcc} 12.2.
All main models were trained for 200 epochs.
Each reported training run used one GPU, and the main reported runs include APS and PDP variants of \method, Single-scale FM, and DiT.
The training batch size was 32 for \method, Single-scale FM, and DiT.
Evaluation and export used batch size 32.
Inference latency was measured on a single NVIDIA H100 80GB HBM3 GPU with CUDA synchronization.

\paragraph{Sampling.}
Sampling uses an explicit Euler solver from \(t=0\) to \(t=1\), following the probability path in Eq.~\eqref{eq:prob_path}.
The default setting uses 20 solver steps at each of the four scales, giving \(\mathrm{NFE}=80\).
For the NFE sweep, iterative generators are evaluated at matched budgets.
For Single-scale FM and DiT, NFE equals the number of sampling steps.

\paragraph{Timing.}
The timing protocol is given in Sec.~\ref{app:latency}.

\section{Scene-Level Confidence Intervals}
\label{app:ci}

Tables~\ref{tab:app_ci_aps} and~\ref{tab:app_ci_pdp} report scene-level bootstrap confidence intervals for APS and PDP metrics.
The intervals summarize variation across scenes rather than across individual spectrum samples. We resample scenes with replacement and recompute the mean metric over the resampled scene set.
The reported interval is given by the 2.5th and 97.5th percentiles of the bootstrap distribution.

\begin{table}[t]
\caption{\textbf{Scene-level confidence intervals for APS metrics under the unified \(\mathrm{NFE}=80\) budget.}
Each cell reports a 95\% confidence interval. Lower values are better.}
\label{tab:app_ci_aps}
\centering
\begingroup
\setlength{\tabcolsep}{4pt}
\renewcommand{\arraystretch}{1.08}
\begin{tabular*}{\linewidth}{@{\extracolsep{\fill}}lcccc@{}}
\toprule
Method & BPL@5 & \(|\Delta\mathrm{ASA}|\) & APS RMSE & APS JSD \\
\midrule
DiT & [5.636, 6.784] & [19.245, 21.985] & [0.042, 0.045] & [0.184, 0.213] \\
Single-scale FM & [2.120, 2.698] & [10.594, 13.011] & [0.019, 0.022] & [0.072, 0.091] \\
\method & [1.781, 2.261] & [9.647, 11.872] & [0.050, 0.051] & [0.064, 0.080] \\
\bottomrule
\end{tabular*}
\endgroup
\end{table}

\begin{table}[t]
\caption{\textbf{Scene-level confidence intervals for PDP metrics under the unified \(\mathrm{NFE}=80\) budget.}
Each cell reports a 95\% confidence interval. Lower values are better.}
\label{tab:app_ci_pdp}
\centering
\begingroup
\setlength{\tabcolsep}{4pt}
\renewcommand{\arraystretch}{1.08}
\begin{tabular*}{\linewidth}{@{\extracolsep{\fill}}lccc@{}}
\toprule
Method & \(|\Delta\mathrm{DS}|\) & PDP RMSE & PDP JSD \\
\midrule
DiT & [32.548, 34.339] & [0.090, 0.096] & [0.221, 0.236] \\
Single-scale FM & [29.000, 30.479] & [0.077, 0.081] & [0.188, 0.202] \\
\method & [28.520, 29.701] & [0.076, 0.080] & [0.180, 0.193] \\
\bottomrule
\end{tabular*}
\endgroup
\end{table}

\section{Qualitative APS Comparisons}
\label{app:qualitative}

\begin{figure}[t]
  \centering
  \includegraphics[width=0.95\linewidth]{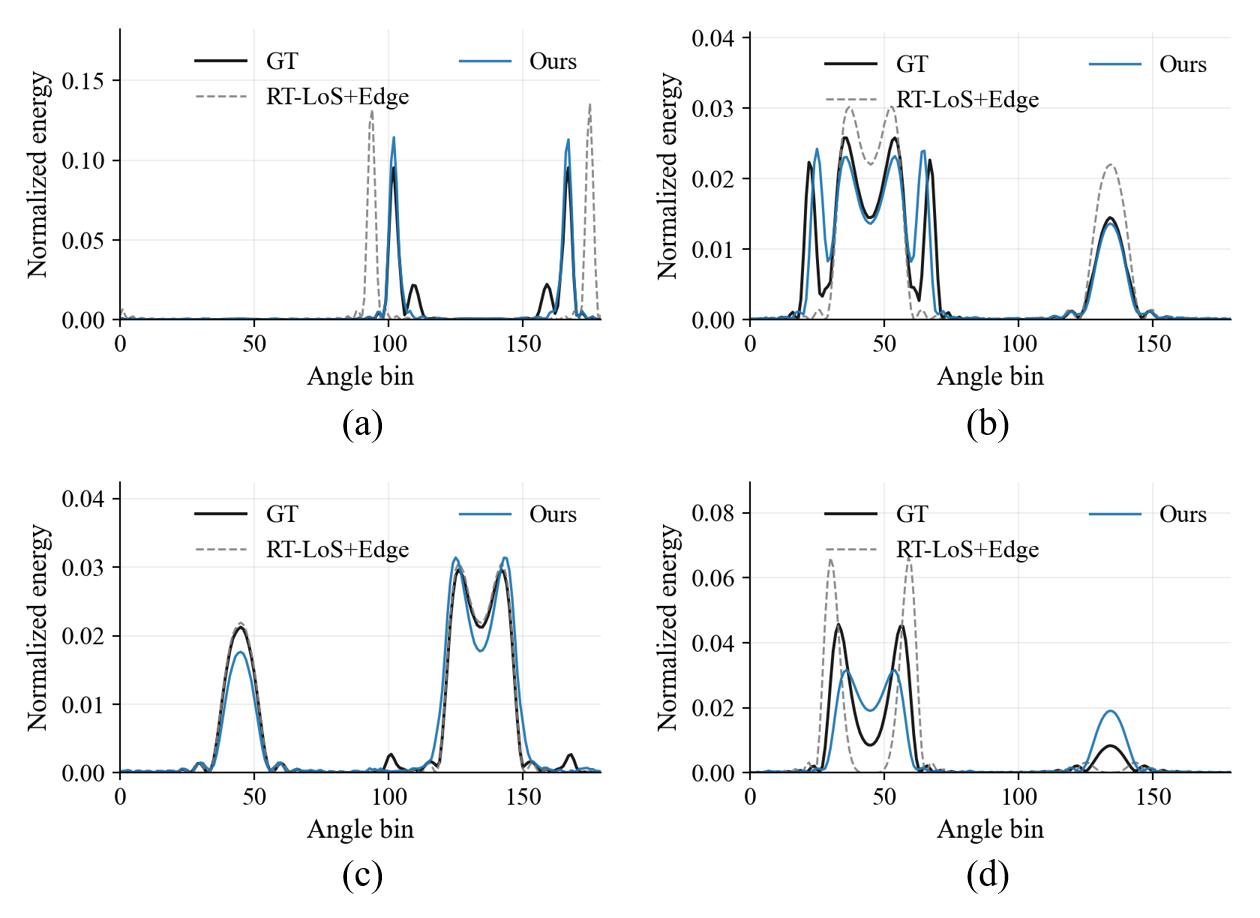}
  \caption{\textbf{Qualitative APS comparisons on representative links.}
  All curves are normalized on the angular grid.
  \method preserves peak locations and relative peak contrast while maintaining secondary lobes and low-power angular mass that affect beam ranking.}
  \label{fig:aps_vis_appendix}
\end{figure}

Figure~\ref{fig:aps_vis_appendix} provides qualitative APS comparisons on representative links.
Each subplot reports the normalized APS over angle bins and compares the ground truth, RT-LoS+Edge, and \method.
The examples illustrate that beam-relevant evaluation depends on dominant directions, peak contrast, and low-power angular mass.

RT-LoS+Edge can be accurate when the angular spectrum is dominated by a small number of strong components.
For more complex links, \method better matches angular locations, peak magnitudes, and fine-grained spectral details.

\section{Baseline Details}
\label{app:baseline_details}

All methods are evaluated on the same packed receiver-resolution target \(y\) with the same validity mask \(m\).
The learned baselines consume the same condition tensor \(I\).

\paragraph{RT-LoS+Edge.}
RT-LoS+Edge retains the LoS path and one dominant roof-edge diffraction path when present.
The reduced path set is mapped to receiver-resolution spectra using the same kernelization, marginalization, packing, and masking pipeline as in Sec.~\ref{sec:method_targets}.
This baseline provides a sparse geometry reference under the same receiver-resolution target interface.

\paragraph{UNet.}
UNet is a single-shot Conv1D regressor that predicts \(\widehat{y}\) in one forward pass.
It is trained with masked regression on valid entries of \(y\), and has \(\mathrm{NFE}=1\).

\paragraph{ChannelGAN.}
ChannelGAN is a conditional GAN that generates \(\widehat{y}\) from \(I\) in one pass.
It is trained with an adversarial objective and a masked reconstruction term on valid entries.
At inference, it has \(\mathrm{NFE}=1\).

\paragraph{DiT.}
DiT follows the diffusion Transformer design~\cite{Peebles2023DiT} and is trained with a noise-prediction objective.
It is sampled through an iterative denoising process.
Each sampling step evaluates the Transformer once, so its NFE equals the number of sampling steps.

\paragraph{Single-scale FM.}
Single-scale FM uses the same packed receiver-resolution target and masked flow-matching objective as \method, but removes the coarse-to-fine autoregressive pyramid.
It generates the full-resolution APS/PDP tensor at one scale.
Its NFE equals the number of solver steps used for the single-scale ODE.
This baseline isolates the effect of coarse-to-fine refinement while keeping the flow-matching formulation fixed.

\section{Full Sampling-Budget Sweep}
\label{app:nfe_sweep_full}

This section provides the full sampling-budget sweep for the iterative generators.
The main text reports decision and dispersion metrics.
Here we additionally report APS/PDP RMSE and JSD under the same matched NFE values.

\begin{table*}[t]
\caption{\textbf{Full APS sampling-budget sweep.}
Each block fixes the sampling budget and compares iterative generators under the same APS metrics.
Lower values are better.}
\label{tab:app_full_nfe_aps}
\centering
\begingroup
\setlength{\tabcolsep}{5pt}
\renewcommand{\arraystretch}{1.08}
\begin{tabular*}{\textwidth}{@{\extracolsep{\fill}}clcccc@{}}
\toprule
NFE & Method & BPL@5 & \(|\Delta\mathrm{ASA}|\) & APS RMSE & APS JSD \\
\midrule
\multirow{3}{*}{12}
& DiT & 4.9231 & 21.9499 & 0.0418 & 0.1732 \\
& Single-scale FM & \textbf{1.7359} & 10.8987 & \textbf{0.0170} & \textbf{0.0644} \\
& \method & 1.8136 & \textbf{10.8908} & 0.0445 & 0.0674 \\
\addlinespace[2pt]
\midrule

\multirow{3}{*}{20}
& DiT & 5.0055 & 18.7989 & 0.0365 & 0.1545 \\
& Single-scale FM & 1.9646 & 11.0663 & \textbf{0.0179} & 0.0681 \\
& \method & \textbf{1.8878} & \textbf{10.6491} & 0.0462 & \textbf{0.0671} \\
\addlinespace[2pt]
\midrule

\multirow{3}{*}{40}
& DiT & 5.4591 & 18.9929 & 0.0391 & 0.1705 \\
& Single-scale FM & 2.1783 & 11.4339 & \textbf{0.0191} & 0.0739 \\
& \method & \textbf{1.9614} & \textbf{10.5938} & 0.0486 & \textbf{0.0694} \\
\addlinespace[2pt]
\midrule

\multirow{3}{*}{60}
& DiT & 5.5973 & 19.1902 & 0.0403 & 0.1778 \\
& Single-scale FM & 2.2656 & 11.2830 & \textbf{0.0196} & 0.0763 \\
& \method & \textbf{2.0393} & \textbf{10.9774} & 0.0496 & \textbf{0.0742} \\
\addlinespace[2pt]
\midrule

\multirow{3}{*}{80}
& DiT & 6.2309 & 20.6737 & 0.0435 & 0.1991 \\
& Single-scale FM & 2.4205 & 11.8679 & \textbf{0.0206} & 0.0822 \\
& \method & \textbf{2.0310} & \textbf{10.8181} & 0.0505 & \textbf{0.0725} \\
\bottomrule
\end{tabular*}
\endgroup
\end{table*}

\begin{table*}[t]
\caption{\textbf{Full PDP sampling-budget sweep.}
Each block fixes the sampling budget and compares iterative generators under the same PDP metrics.
Lower values are better.}
\label{tab:app_full_nfe_pdp}
\centering
\begingroup
\setlength{\tabcolsep}{6pt}
\renewcommand{\arraystretch}{1.08}
\begin{tabular*}{\textwidth}{@{\extracolsep{\fill}}clccc@{}}
\toprule
NFE & Method & \(|\Delta\mathrm{DS}|\) & PDP RMSE & PDP JSD \\
\midrule
\multirow{3}{*}{12}
& DiT & 44.0734 & 0.0911 & 0.2090 \\
& Single-scale FM & 31.4224 & \textbf{0.0757} & 0.1810 \\
& \method & \textbf{28.6115} & 0.0760 & \textbf{0.1755} \\
\addlinespace[2pt]
\midrule

\multirow{3}{*}{20}
& DiT & 36.2404 & 0.0908 & 0.2188 \\
& Single-scale FM & 30.1083 & \textbf{0.0773} & 0.1896 \\
& \method & \textbf{29.1227} & 0.0778 & \textbf{0.1867} \\
\addlinespace[2pt]
\midrule

\multirow{3}{*}{40}
& DiT & 33.6613 & 0.0915 & 0.2252 \\
& Single-scale FM & 29.7640 & \textbf{0.0786} & \textbf{0.1933} \\
& \method & \textbf{29.4816} & 0.0792 & 0.1938 \\
\addlinespace[2pt]
\midrule

\multirow{3}{*}{60}
& DiT & 33.5463 & 0.0922 & 0.2272 \\
& Single-scale FM & 29.7465 & \textbf{0.0790} & 0.1944 \\
& \method & \textbf{29.4881} & 0.0797 & \textbf{0.1941} \\
\addlinespace[2pt]
\midrule

\multirow{3}{*}{80}
& DiT & 33.4298 & 0.0929 & 0.2287 \\
& Single-scale FM & 29.7551 & 0.0792 & 0.1952 \\
& \method & \textbf{29.1227} & \textbf{0.0778} & \textbf{0.1867} \\
\bottomrule
\end{tabular*}
\endgroup
\end{table*}

\section{Timing Protocol}
\label{app:latency}

Runtime is measured for model inference with GPU synchronization.
Data loading, disk I/O, and host-device transfer are excluded.
All iterative generators are timed under the unified \(\mathrm{NFE}=80\) budget unless otherwise stated.
NFE matches the number of model evaluations, but the runtime of one evaluation can differ across architectures.
Therefore, we report both NFE-controlled accuracy metrics and measured runtime.

For \method, early evaluations are performed on coarser APS/PDP grids before full-resolution refinement.
Single-scale FM evaluates the full-resolution spectrum at every step.
DiT follows a single-scale diffusion implementation.
These details clarify the timing protocol and should not be interpreted as a general complexity ranking among model families.

\section{Broader Impacts}
\label{app:broader_impacts}

This work aims to improve reproducible evaluation of receiver-resolution wireless spectrum maps.
Potential positive impacts include more reliable beam-management studies, standardized APS/PDP learning targets, and efficient dense spectrum queries for wireless-network research.
Potential negative impacts may arise if radio-environment inference tools are used for unauthorized monitoring, sensitive infrastructure analysis, or deployment decisions without appropriate validation.
The benchmark should therefore be released with clear intended-use documentation and should not be used to infer private user behavior or sensitive deployment information.

\end{document}